\documentclass[11pt,a4paper]{article}
\usepackage[hyperref]{naaclhlt2019}
\usepackage{times}
\usepackage{latexsym}

\usepackage{url}

\usepackage{amsfonts}
\usepackage{amsmath}
\usepackage{graphicx}

\usepackage{color}

\usepackage{enumitem}
\usepackage{linguex}

\aclfinalcopy % Uncomment this line for the final submission
\title{Grounded Agreement Games: Emphasizing\\ Conversational
  Grounding in Visual Dialogue Settings}

\author{David Schlangen\\
  Computational Linguistics / Department of Linguistics\\
  University of Potsdam, Germany\\
  {\tt david.schlangen@uni-potsdam.de} \\}

\date{}

\begin{document}
\maketitle
\begin{abstract}
  % After many years on the sidelines, \emph{dialogue} as a topic in computational linguistics is enjoying time on the main playing field again. Unlike the early work, however, which
Where early work on dialogue in Computational Linguistics
  put much emphasis on dialogue structure and its relation to the mental states of the dialogue participants (e.g., Allen 1979, Grosz \& Sidner 1986), current work mostly % limits itself to dealing with flat sequences of utterances, reducing
  reduces dialogue to the task of producing at any one time a next utterance; e.g.\ in \emph{neural chatbot} or \emph{Visual Dialogue} settings.
  As a methodological decision, this is sound: Even the longest journey is a sequence of steps. %, as it were.
  It becomes detrimental, however, when the tasks and datasets from which dialogue behaviour is to be learned
  % and the settings that are used to create them
  are tailored too much to this framing of the problem. %, and when dialogue is treated as a simple scaling up from modelling single utterance/response pairs.
  In this short note, we % argue that this is the case for Visual Dialogue, where the setting does not provide a structuring purpose, which is otherwise naturally found in dialogue. We
  describe a family of settings which still allow to keep dialogues simple, but add a constraint that makes parti\-cipants care about reaching mutual understanding. In such \emph{agreement games}, there is a secondary, but explicit goal besides the task level goal, and that is to reach mutual understanding about whether the task level goal has been reached. As we argue, this naturally triggers meta-semantic interaction and mutual engagement, and hence leads to richer data from which to induce models.
%  , potentially, models that handle more than local coherence can be learned.
%  As we argue in this short note, there is a way to imbue these currently popular settings with a structuring purpose that has so far been lacking.   In an \emph{agreement game} setting, participants need to explictly reach an agreement on the status of their dialogue. As we show, this naturally triggers meta-semantic interaction and hence leads to data from which, potentially, models that handle more than local coherence can be learned.
\end{abstract}

\section{Introduction}
\label{sec:intro}

\nocite{allen-1979-plans,gs:aisd}

If you're good at replying to a single request, are you also likely to be good at doing dialogue? Much current work seems to assume that the answer to this question is yes, in that it attempts a scaling up from single pairs of utterance plus response  to longer dialogues: See, e.g., the work on \emph{neural chatbots} following on from \cite{Serban2016}, where the main evaluation metric is ``next utterance retrieval''; and on \emph{visual dialogue} \cite{visdial}, which views itself as a natural extension of \emph{visual question answering} \cite{VQA2015}.

If you assume, however, that dialogue crucially is a \emph{joint project} between its participants in a way that single exchanges are not, you're likely to put more focus on \emph{coordination phenomena} \cite{clark:ul}, but may end up with settings that combine multiple language capabilities in ways that current methods cannot yet seem to capture. (See, for example, the dialogues collected in \cite{pentoref2016}.) In this short paper, we contribute a type of setting that introduces such coordination phenomena, while still allowing for control of the complexity of the resulting interaction.\footnote{%
  The datasets discussed in this paper can be viewed in a common format using the code at \url{https://github.com/clp-research/sempix}, \cite{schlangen:iwcs19}.
}

%\clearpage
\section{Visual Dialogue as Example of the Scaling Up Approach}
\label{sec:up}

\begin{figure}
  \begin{center}
   \hspace*{-.7cm}
 \includegraphics[width=1.1\linewidth]{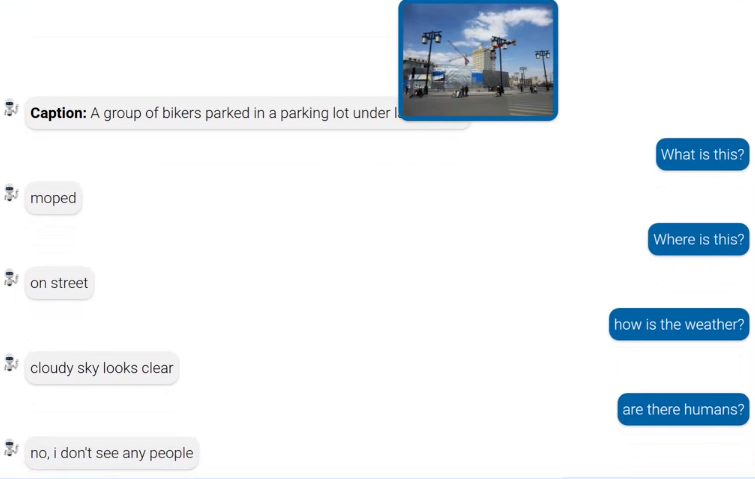}
 \caption{Original example of Visual Dialogue System, from official demo video ({\small \url{https://vimeo.com/193092429}}) }
 \label{fig:vimeo}
\end{center}
%\vspace*{-.5cm}
\end{figure}

Figure~\ref{fig:vimeo} shows an example interaction with the original Visual Dialogue system \cite{visdial}. The competence of the system is impressive from a multimodal grounding perspective --- it gets right several questions aiming at different aspects of the image. It is also clear, however, that this is a modest step beyond single-shot visual question answering \cite{VQA2015}. It seems that here 
% While impressive from a multimodal grounding perspective, this excerpt is less so from a perspective of dialogue modelling. It is clear that here
the (human) questioner is doing all the work of keeping the dialogue alive, and there is little that suggest that the answerer is keeping any state about the dialogue. Later work by \citet{Kottur2018} on the ``visual dialogue'' dataset \cite{visdial} indeed identified \emph{co-reference} in the questions as the main issue that distinguishes this setting from one-shot question answering.
%, hence adding a single phenomenon only to what still fundamentally is a sequence of questions and answers.

One shortcoming of this setting---that the questioner was not provided with a good reason for \emph{why} they are asking questions in the first place---was addressed in some related work: In the \emph{GuessWhat?}\ setting  introduced by \citet{DeVries2016:guesswhat} at around the same time as Visual Dialogue, a questioner is asking polar questions about an image, with the goal of identifying an object known only to the questioner. In the ALICE variant of the visual dialogue setting, the questioner is asking questions with the later goal in mind of identifying the image from a set also containing distractor images \cite{Chattopadhyay2017}.\footnote{%
  Interestingly, this setup was created to evaluate dialogue agents trained on the visual dialogue dataset, not to collect another type of data.}
These variants of the general setting provide purpose to the questioner, but not to the answerer, which is the target of the modelling effort; and, crucially, it does not give the dialogue a \emph{joint purpose}, a shared sense of semantic ownership of the interaction, which is a central feature of most genres of human interaction \cite{clark:ul}.

\begin{figure}[ht]
\begin{center}
 \includegraphics[width=0.65\linewidth]{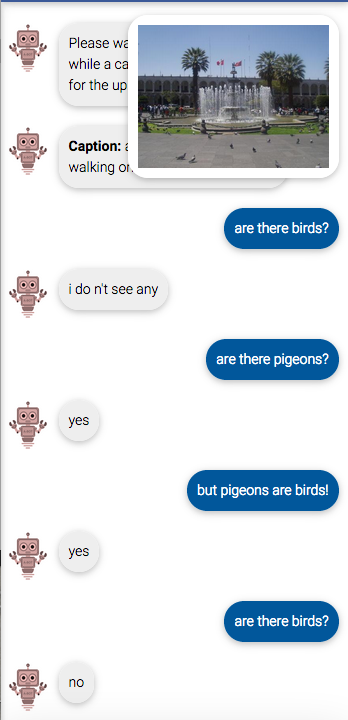}
 \caption{Interacting with a visual chat bot, see \url{http://demo.visualdialog.org/}}
 \label{fig:pigeons}
\end{center}
%\vspace*{-1cm}
\end{figure}

Coming back to the visual dialogue setting, it can be assumed that the crowd workers that created the original data did try to orient themselves to the usual maxims that govern conversational behaviour. However, being constrained by the rigid roles of questioner and answerer, and with the perceptual task being so easy for them,
% This is by design: The setting in which the data was created from which such ``visual dialogue agents'' are to be learned was designed to make clear to the participants the limits of their respective roles. A presumably unplanned consequence of the setting, however, is that since the perceptual task is so easy for humans,
a need for dealing with miscommunication never arose for them and hence no such strategies can be learned from that data. That this is missing from the resulting agents can easily been shown in cases where something goes wrong, but normal repair mechanisms \cite[see e.g.][]{hayashi2013conversational} are not available, as in the example interaction we created shown in Figure~\ref{fig:pigeons}.\footnote{%
  As the authors have made the laudable effort of opening their system to the general public, one can find similar examples of interactions that laypeople had with the agent and posted on social media; e.g. \url{https://twitter.com/r_speer/status/1037358574735904768}.
}

\section{Agreement Games}
\label{sec:agree}

In Herbert Clark's (\citeyear{clark:ul}) model of dialogue,\footnote{%
  Bringing together ideas from Conversation Analysis \cite{sacksetal:turnt_full,schegloff:sequencebook} and also the design of Human/Computer interaction \cite{norman:design}.

} the mutual need for ensuring understanding---``sufficient to current purposes''---is the main structuring force in dialogue. As a metaphor for this interaction management process, Clark uses the notion of a ``secondary track'', on which the constant negotiation of this understanding happens. This can be done through quite subtle methods, such as simply producing a continuation that displays through its fit an understanding of the previous utterance, as well as through specially designed markers (such as feedback utterances like ``uhu''). This model has been influential in the design of spoken dialogue systems \cite{traum:commodgroun,skantze:explspecom}, where it has been incorporated in the design of dialogue state update rules. In the data-driven era, however, it seems to have become less well known, and many datasets almost appear as to be designed in such a way as to limit the possibility of grounding interactions.
%likely because of the variety of markers that are thought to express it.

The idea behind our setting of ``agreement games'' is to make this secondary track more prominent and hence more easy to pick up from the data,
% . The idea is as follows: In an agreement game,
by making reaching mutual understanding on the answer to the game question %becomes
an explicit goal. Or, in Clark's term, the \emph{grounding criterion} for answering the question is raised so as to make reaching mutual understanding on it an explicit, rather than as normally implicit, goal.

The representational challenge is that it is an abstract object---the understanding of the discourse---that is jointly constructed, purely through verbal actions.

\paragraph{More formally}
\label{sec:formally}

An \emph{Agreement Game} is a dialogue game with two regular participants, $\mathcal{P} =  \{P_1, P_2\}$, and a disinterested third participant, $N$ (for \emph{Nature}). $N$ poses a question $Q$ to the players $\mathcal{P}$, and provides them with information $I$ required to answer the question; possibly split up over the players. If $I$ contains visual information, we call the game a \emph{Grounded} Agreement Game.

The players can exchange messages in an unrestricted way. The game ends when one of the players explicitly proposes an answer $A$ and the other player explicity agrees with the proposal. As the answer $A$ will be based on a construal of $I$, the agreement on $A$ is also an agreement on that construal. Optionally, a reward can be given to the players after they have provided their joint answer, tied to some measure of quality of $A$.

\hspace*{.5\baselineskip}

We illustrate the concept by discussing some instantiations that we have recently experimented with.\footnote{%
  Concurrently, two other datasets have recently been published using settings that can also be viewed as grounded agreement games: In the PhotoBook setting \cite{Haber2019}, two players must agree on which images from sets of (natural) images that they invidivdually are presented with are identical. They do this repeatedly, so that occasion for re-references occurs. In the ``One in Common'' setting \cite{Udagawa2019}, players are presented with images of overlapping, but non-congruent views on a synthetic scene, and their goal is to determine which objects belong to the overlapping part.
}

%\footnote{%
%   We have so far realised variants of this game type using a tool that facilitates written interaction \cite{slurk.semdial18}, where we make it a condition of the game that it ends by the written acknowledgement (through special commands) of both players that they agree with the provided answer; it should also be possible to realise this constraint in spoken interaction. 
% }
% $^,$\footnote{
%   \das{To be added} Many familiar dialogue games from the pycholinguistics literature (Krauss \& Weinheimer, etc.., MapTask) can be understood as being agreement games. But not tailored towards producing data for machine learning.. Also, recently introduced PhotoBook dataset.
% }

\section{Some Examples}
\label{sec:ex}

\subsection{The MeetUp Game}
\label{sec:meetup}

In the MeetUp game \cite{schlangen:aix,ilieatal:meetup19},\footnote{%
  \url{https://github.com/clp-research/meetup}
} the two participants are presented with an environment through which they can (separately) navigate and which is represented to static photographs or real indoor scenes (e.g., a picture of a bedroom, a kitchen, etc.) Their goal is to meet up in the same room, of a type previously told to them. (E.g., they might be told at the start of the game: ``your goal is to meet up in a room of type \emph{kitchen}.'') As the positions of the players are not represented graphically, the only way they can be sure of whether they have reached that goal is by conversing (via chat messages). Once they have come to the conclusion that they are in the same room, they can end the game by each sending a special signal. If they are indeed in such a winning constellation, they will receive a bonus.

Unlike the Visual Dialogue setting discussed above, this setting ensures informational symmetry between the participants (both have access to the same \emph{type} of information; but not the same information, as they can't ``see'' each other). More importantly, however, the constraint that the game only ends if they both agree ensures a ``committment symmetry'', where the success of the game must be ensured by both participants. The design also provides for a clear ``relevance place'' at which an opportunity arises for \emph{semantic negotiation}, namely, before the final decision is made.\footnote{%
  ``Relevance place'' in analogy to the \emph{transition relevance places} studied for turn taking in conversational analysis by \citet{sacksetal:turnt_full}.}
An example of this is shown in the example below. (The number in the parentheses indicate the time, relative to the beginning of the interaction, when the utterance was made.)

\ex. \label{ex:utility}
%\a. 
B (00:00:34): okay I think I'm there if I understand utility room 
\\ %\b.
B (00:00:42): It has a washer and dryer 
\\ %\b.
A (00:00:46): I was wondering too. This is sorta like a laundry room. 
\\ %\b.
A (00:00:55): This has pet bowl on the floor below a window. 
\\ %\b.
B (00:01:00): ok... let us keep looking 
\\ %\b.
A (00:01:22): And a small kids looking suit hanging on the wall. And a big banner above the window. 
\\ %\b.
B (00:01:33): Are you saying a utility room is like a laundry room?
\\ %\b.
B (00:02:00): let me find you 
\\ %\b.
A (00:02:07): Google says, a room equipped with appliances for washing and other domestic work.
\\ %\b.
A (00:02:09): So I think so.

\subsection{The MatchIt Game}
\label{sec:matchit}

The MatchIt Game (Ilinykh et al., forthcoming) is a yet further simplified visual game. Here, the goal simply is to decide whether you and your partner are both looking at the same image (of the same genre as in MeetUp). In that sense, it is a reduction of the MeetUP game to the final stage, taking out the navigation aspect. As example \ref{ex:kitchenbath} shows, this can similarly lead to meta-semantic interaction, where classifications are revised. As \ref{ex:bike} shows, even in cases where a decision can be reached quickly, there can be an explicit mutual confirmation step, before the (silent) decision signal is sent.

\ex. \label{ex:kitchenbath}
%\a.
B (00:00:25): white kitchen?
\\ %\b.
A (00:00:25): im in a bathroom
\\ %\b.
B (00:00:28): ah
\\ %\b.
B (00:00:32): well wait
\\ %\b.
B (00:00:38): there is something that looks like a big bath
\\ %\b.
B (00:00:44): is it all white?
\\ %\b.
A (00:00:54): yes its white and I see a bit of a kitchen
\\ %\b.
A (00:01:11): yes
\\ %\b.
B (00:01:11): are you sure it's a bathroom lol
\\ %\b.
A (00:01:16): no its not a bathroom haha

\ex. \label{ex:bike}
%\a.
A (00:00:24): i see stairs
\\ %\b.
B (00:00:25): I see a staircase with a bike with wicker basket at the bottom of the staircase
\\ %\b.
B (00:00:31): do you have a bike?
\\ %\b.
A (00:00:39): no bike
\\ %\b.
B: (00:00:46): okay..it is different
\\ %\b.
A (00:00:54): yes

\subsection{The Concept Learning Game}
\label{sec:concept}

A third setting that we have explored \cite{attari:sigdial19} brings conceptual negotiation more clearly into the foreground. In that game, the players are presented with images of birds of particular species and are tasked with coming up with a description of common properties. Again, the final answer has to be approved by both participants. As \ref{ex:orange} shows, this can lead to an explicit negotiation of conceptual content.

\ex. \label{ex:orange}
\hspace*{-4ex}
\begin{tabular}[t]{l@{$\;$}p{.95\linewidth}}
  A: & Looks like the birds under 2 have red-orange feet. \\
  B: & The difference that I notice is that the birds in Section 1 are light feathered vs. the dark feathered birds of Section 2. \\
  A: & Ah, I like your answer better than mine.\\ 
  B: & \emph{/answer} The birds in section 1 do not have red-orange feet like the birds in section 2. Also, the feathers of the birds in Section 1 are light-colored vs. the dark-colored feathers of the birds in Section 2. \\
  A: & \emph{/agree}
\end{tabular}

\section{Conclusions}
\label{sec:conc}

We have argued that some prominent current dialogue settings lack room for the occurence of \emph{coordination phenomena} prevalent in %, and indeed defining of,
natural dialogue. We have shown a simple condition that brings out the need for coordination in an explicit way, but still can easily be added to controlled (and controllable) dialogue settings.

%The tasks that we have described here are informationally symmetric; this however is not a requirement, and tasks like visual dialogue, where only one player has an informational need and makes a task-level decision, can still be used, as long as the dialogue / understanding level decision is put to all participants.

%\clearpage
\bibliographystyle{acl_natbib}
\bibliography{/Users/das/work/projects/MyDocuments/BibTeX/all-lit.bib}

\begin{thebibliography}{22}
\expandafter\ifx\csname natexlab\endcsname\relax\def\natexlab#1{#1}\fi

\bibitem[{Allen(1979)}]{allen-1979-plans}
James~F. Allen. 1979.
\newblock \href {https://doi.org/10.3115/982163.982187} {Plans, inference, and
  indirect speech acts}.
\newblock In \emph{Proceedings of the 17th Annual Meeting of the Association
  for Computational Linguistics}, pages 85--87, La Jolla, California, USA.
  Association for Computational Linguistics.

\bibitem[{Antol et~al.(2015)Antol, Agrawal, Lu, Mitchell, Batra, Zitnick, and
  Parikh}]{VQA2015}
Stanislaw Antol, Aishwarya Agrawal, Jiasen Lu, Margaret Mitchell, Dhruv Batra,
  C.~Lawrence Zitnick, and Devi Parikh. 2015.
\newblock Vqa: Visual question answering.
\newblock In \emph{International Conference on Computer Vision (ICCV)}.

\bibitem[{Attari et~al.(2019)Attari, Heckmann, and
  Schlangen}]{attari:sigdial19}
Nazia Attari, Martin Heckmann, and David Schlangen. 2019.
\newblock From explainability to explanation: Using a dialogue setting to
  elicit annotations with justifications.
\newblock In \emph{Proceedings of SIGdial 2019}, Stockholm, Sweden.

\bibitem[{Chattopadhyay et~al.(2017)Chattopadhyay, Yadav, Prabhu,
  Chandrasekaran, Das, Lee, Batra, and Parikh}]{Chattopadhyay2017}
Prithvijit Chattopadhyay, Deshraj Yadav, Viraj Prabhu, Arjun Chandrasekaran,
  Abhishek Das, Stefan Lee, Dhruv Batra, and Devi Parikh. 2017.
\newblock \href {http://arxiv.org/abs/1708.05122} {{Evaluating Visual
  Conversational Agents via Cooperative Human-AI Games}}.
\newblock \emph{ArXiv}.

\bibitem[{Clark(1996)}]{clark:ul}
Herbert~H. Clark. 1996.
\newblock \emph{Using Language}.
\newblock Cambridge University Press, Cambridge.

\bibitem[{Das et~al.(2017)Das, Kottur, Gupta, Singh, Yadav, Moura, Parikh, and
  Batra}]{visdial}
Abhishek Das, Satwik Kottur, Khushi Gupta, Avi Singh, Deshraj Yadav,
  Jos\'e~M.F. Moura, Devi Parikh, and Dhruv Batra. 2017.
\newblock {V}isual {D}ialog.
\newblock In \emph{Proceedings of the IEEE Conference on Computer Vision and
  Pattern Recognition (CVPR)}.

\bibitem[{{De Vries} et~al.(2017){De Vries}, Strub, Chandar, Pietquin,
  Larochelle, and Courville}]{DeVries2016:guesswhat}
Harm {De Vries}, Florian Strub, Sarath Chandar, Olivier Pietquin, Hugo
  Larochelle, and Aaron Courville. 2017.
\newblock {GuessWhat?! Visual object discovery through multi-modal dialogue}.
\newblock In \emph{CVPR 2017}.

\bibitem[{Grosz and Sidner(1986)}]{gs:aisd}
Barbara~J. Grosz and Candace~L. Sidner. 1986.
\newblock Attention, intentions, and the structure of discourse.
\newblock \emph{Computational Linguistics}, 12(3):175--204.

\bibitem[{Haber et~al.(2019)Haber, Baumg{\"{a}}rtner, Takmaz, Gelderloos,
  Bruni, and Fern{\'{a}}ndez}]{Haber2019}
Janosch Haber, Tim Baumg{\"{a}}rtner, Ece Takmaz, Lieke Gelderloos, Elia Bruni,
  and Raquel Fern{\'{a}}ndez. 2019.
\newblock {The PhotoBook Dataset: Building Common Ground through
  Visually-Grounded Dialogue}.
\newblock In \emph{Proceedings of the 2019 meeting of the Association for
  Computational Linguistics}, Florence, Italy.

\bibitem[{Hayashi et~al.(2013)Hayashi, Raymond, and
  Sidnell}]{hayashi2013conversational}
Makoto Hayashi, Geoffrey Raymond, and Jack Sidnell. 2013.
\newblock \emph{Conversational Repair and Human Understanding}.
\newblock OAPEN Library. Cambridge University Press.

\bibitem[{Ilinykh et~al.(2019)Ilinykh, Zarrie{\ss}, and
  Schlangen}]{ilieatal:meetup19}
Nikolai Ilinykh, Sina Zarrie{\ss}, and David Schlangen. 2019.
\newblock Meetup! a corpus of joint activity dialogues in a visual environment.
\newblock In \emph{Proceedings of the 23rd Workshop on the Semantics and
  Pragmatics of Dialogue (SemDial 2019 / LondonLogue)}, London, UK.

\bibitem[{Kottur et~al.(2018)Kottur, {F Moura}, Parikh, Batra, and
  Rohrbach}]{Kottur2018}
Satwik Kottur, Jos{\'{e}}~M {F Moura}, Devi Parikh, Dhruv Batra, and Marcus
  Rohrbach. 2018.
\newblock \href {http://arxiv.org/abs/1809.01816} {{Visual Coreference
  Resolution in Visual Dialog using Neural Module Networks}}.
\newblock \emph{ArXiv}.

\bibitem[{Norman(1988)}]{norman:design}
Donald~A. Norman. 1988.
\newblock \emph{The Design of Everyday Things}.
\newblock Basic Books, Inc., New York, NY, USA.

\bibitem[{Sacks et~al.(1974)Sacks, Schegloff, and
  Jefferson}]{sacksetal:turnt_full}
Harvey Sacks, Emanuel~A. Schegloff, and Gail~A. Jefferson. 1974.
\newblock A simplest systematic for the organization of turn-taking in
  conversation.
\newblock \emph{Language}, 50:735--996.

\bibitem[{Schegloff(2007)}]{schegloff:sequencebook}
Emanuel~A. Schegloff. 2007.
\newblock \emph{Sequence Organization in Interaction}.
\newblock Cambridge University Press, Cambridge, UK.

\bibitem[{Schlangen(2019)}]{schlangen:iwcs19}
David Schlangen. 2019.
\newblock Natural language semantics with pictures: Some language \& vision
  datasets and potential uses for computational semantics.
\newblock In \emph{Proceedings of the International Conference on Computational
  Semantics (IWCS)}, Gothenburg.

\bibitem[{Schlangen et~al.(2018)Schlangen, Ilinykh, and
  Zarrieß}]{schlangen:aix}
David Schlangen, Nikolai Ilinykh, and Sina Zarrieß. 2018.
\newblock {MeetUp! A Task For Modelling Visual Dialogue}.
\newblock In \emph{Short Paper Proceedings of the 22nd Workshop on the
  Semantics and Pragmatics of Dialogue (AixDial / semdial 2018)}.

\bibitem[{Serban et~al.(2016)Serban, Sordoni, Bengio, Courville, and
  Pineau}]{Serban2016}
Iulian~V. Serban, Alessandro Sordoni, Yoshua Bengio, Aaron Courville, and
  Joelle Pineau. 2016.
\newblock {Building End-To-End Dialogue Systems Using Generative Hierarchical
  Neural Network Models}.
\newblock In \emph{Proceedings of the Meeting of the Association for the
  Advancement of Artificial Intelligence (AAAI)}.

\bibitem[{Skantze(2005)}]{skantze:explspecom}
Gabriel Skantze. 2005.
\newblock Exploring human error recovery strategies: Implications for spoken
  dialogue systems.
\newblock \emph{Speech Communication}, 45(3):325--341.

\bibitem[{Traum(1999)}]{traum:commodgroun}
David~R. Traum. 1999.
\newblock Computational models of grounding in collaborative systems.
\newblock In \emph{Proceedings of the AAAI Fall Symposium on psychological
  models of communication}, pages 124--131.

\bibitem[{Udagawa and Aizawa(2019)}]{Udagawa2019}
Takuma Udagawa and Akiko Aizawa. 2019.
\newblock \href {http://arxiv.org/abs/arXiv:1907.03399v1} {{A Natural Language
  Corpus of Common Grounding under Continuous and Partially-Observable
  Context}}.
\newblock In \emph{Proceedings of AAAI}.

\bibitem[{Zarrie{\ss} et~al.(2016)Zarrie{\ss}, Hough, Kennington,
  Manuvinakurike, DeVault, Fern{\'a}ndez, and Schlangen}]{pentoref2016}
Sina Zarrie{\ss}, Julian Hough, Casey Kennington, Rames Manuvinakurike, David
  DeVault, Raquel Fern{\'a}ndez, and David Schlangen. 2016.
\newblock Pentoref: A corpus of spoken references in task-oriented dialogues.
\newblock In \emph{Proceedings of LREC 2016}, Portoroz, Slovenia.

\end{thebibliography}

\end{document}